# Digitization of Document and Information Extraction using OCR

Rasha Sinha
*Department of Information Science and Engineering*
*RV College of Engineering*
*Bangalore, India*
rashasinha.is21@rvce.edu.in

Mrs. Rekha B S
*Assistant Professor*
*Department of Information Science and Engineering*
*RV College of Engineering*
*Bangalore, India*
rekhabs@rvce.edu.in

*Abstract*— **Retrieving accurate details from documents is a crucial task, especially when handling a combination of scanned images and native digital formats. This document presents a combined framework for text extraction that merges Optical Character Recognition (OCR) techniques with Large Language Models (LLMs) to deliver structured outputs enriched by contextual understanding and confidence indicators. Scanned files are processed using OCR engines, while digital files are interpreted through layout-aware libraries. The extracted raw text is subsequently analyzed by an LLM to identify key-value pairs and resolve ambiguities. A comparative analysis of different OCR tools is presented to evaluate their effectiveness concerning accuracy, layout recognition, and processing speed. The approach demonstrates significant improvements over traditional rule-based and template-based methods, offering enhanced flexibility and semantic precision across different document categories**

I. INTRODUCTION

In today's data-focused environment, a considerable volume of essential information exists in unstructured document formats like scanned images, PDFs, and original digital files. Extracting organized information from these varied document sources poses a complicated challenge because of layout inconsistencies, embedded text elements, and the decline in quality in scans. Even though digital documents may preserve certain types of structural metadata, their differing formats frequently make direct parsing difficult. Conversely, scanned documents present extra challenges like noise, distortion, and indistinct text boundaries that obstruct precise recognition and understanding. Conventional rule-based or template-specific extraction methods frequently struggle when there are these diverse input. To solve these challenges, this study introduces a hybrid extraction framework that combines traditional Optical Character Recognition (OCR) techniques with Large Language Models (LLMs). The method utilizes OCR software to analyze scanned files, along with layout-aware parsing libraries for digital formats, and then employs LLMs for semantic interpretation to obtain significant key-value pairs and contextual details.

The primary contributions of this study consist of the following:

- This research performs a systematic evaluation of Tesseract, DocTR, and Google Vision API across various document types, emphasizing their comparative accuracy, layout retention, and practical challenges
- A cohesive pipeline is suggested for managing scanned and native digital documents, incorporating preprocessing techniques such as binarization, skew correction, and segmentation to improve text extraction and structural uniformity.
- This research utilizes extensive language models to extract meaningful data from unstructured inputs, with outputs structured and annotated with confidence scores to support accurate downstream processing.

This framework showcases a scalable approach for document comprehension in business and research scenarios, surpassing traditional extraction systems in terms of flexibility and precision.

II. LITERATURE REVIEW

Digitization and Processing of Documentary Information: This document thoroughly examines the complete procedure of transforming physical documents into superior digital formats, emphasizing scanning, PDF processing, and optical character recognition (OCR) methods. It analyzes how essential elements like image resolution and compression techniques affect the sharpness and precision of digitized material. The research additionally examines software tools such as Adobe Acrobat and CAJViewer, analyzing their effectiveness in improving document readability, searchability, and storage efficiency. The paper offers insightful recommendations for enhancing the preservation, accessibility, and handling of documentary information in different archival and administrative contexts by outlining best practices and technical factors. [1]

A Study of The OCR Development History and Directions of Development: This document offers an extensive overview of the evolution of Optical Character Recognition (OCR) technology, following its advancement from initial mechanical and template-driven techniques in the 1950s to contemporary cutting-edge methods utilizing deep learning. It outlines important technological advancements, such as the implementation of machine learning algorithms and, more recently, convolutional neural networks (CNNs) and recurrent neural networks (RNNs), which have substantially improved OCR precision and reliability. The research systematically classifies OCR methods into conventional approaches such as template matching and structural analysis, alongside contemporary neural network models, offering an essential evaluation of their efficiency across various text types, including both printed and handwritten materials. Additionally, the study tackles difficulties in OCR, including the recognition of cursive writing and handling low-quality or noisy pictures, and assesses how these problems have been reduced over time. It categorizes OCR research according to language groups and dataset benchmarks, emphasizing the growing emphasis on multilingual OCR systems and the application of standardized datasets like CEDAR and MNIST for assessment. Ultimately, the paper addresses the future pathways of OCR, highlighting the incorporation of AI-powered methods for improved digitization, natural language processing, and practical applications. [2]

Optical Character Recognition from Text Image: This paper proposes an OCR method that enhances recognition accuracy by focusing on the extraction of distinctive texture and topological features of characters, such as corner points and area ratios. Instead of relying solely on pixel-based or statistical models, the approach analyzes the geometric and structural properties of each character to create robust feature representations. These features are then matched against a set of trained templates for precise identification. The method is particularly effective in handling diverse fonts, varying handwriting styles, and degraded or noisy images, which commonly challenge traditional OCR systems. As a result, the technique significantly improves text extraction quality in scanned documents, handwritten notes, and typewritten materials, making it valuable for applications that demand high accuracy in document digitization. [3]

Confidence in the Reasoning of Large Language Models: This research assesses the confidence exhibited by large language models (LLMs) like GPT4o, GPT4-turbo, and Mistral in their responses and examines how this confidence correlates with accuracy. Confidence can be assessed qualitatively, based on whether the model maintains its original answer when asked to rethink, and quantitatively, through self-reported confidence ratings. Evaluated on tough benchmarks related to causal reasoning, logical fallacies, and statistical dilemmas, the LLMs perform better than random guesses but frequently alter their responses upon reflection, with subsequent answers occasionally being less precise. Although greater confidence typically aligns with improved accuracy, LLMs exhibit a significant propensity to exaggerate their certainty and demonstrate an inconsistent grasp of uncertainty. These results emphasize that although current LLMs demonstrate remarkable reasoning abilities, they lack true introspective awareness of their confidence, which calls for caution in understanding their assertive replies. [4]

Digitization of documents using OCR: The suggested system intends to automate the assessment and grading of handwritten exam papers by utilizing Optical Character Recognition (OCR) and Natural Language Processing (NLP) methods. Handwritten answer sheets are scanned and undergo pre-processing to improve image quality, subsequently followed by text extraction through OCR (notably PyTesseract). The retrieved text is subsequently examined by correlating keywords and their synonyms from a curated database utilizing NLP tools such as NLTK and WordNet to evaluate the accuracy of answers. The system computes scores according to keyword relevance and similarity metrics, guaranteeing a consistent and unbiased assessment while minimizing manual work. This method simplifies the evaluation procedure, offering automatic result creation and convenient oversight through an admin dashboard. [5]

### III. BACKGROUND AND RELATED WORKS
*A. Classic OCR Techniques*

Optical Character Recognition (OCR) serves as the foundation for transforming handwritten exam sheets into a digital format. Conventional OCR tools like Tesseract, EasyOCR, and Google Vision API are commonly employed in tasks involving document digitization. Among them, Tesseract is an open-source tool that accommodates various languages; nonetheless, it can have difficulty with intricate layouts and low-quality handwriting. EasyOCR provides enhanced support for cursive and handwritten writing, but it has restricted customization options and somewhat lower accuracy in parsing structured documents. The Google Vision API, a paid service, provides high precision and strong language identification but includes related usage fees. Although these OCR tools offer fundamental text extraction functions, their efficiency differs considerably based on language variety, image quality, and the structural arrangement of the source documents.

*B. Parsers for Digital Documents*

For digital files like PDFs and Word documents, tools like PyMuPDF, pdfplumber, and python-docx provide programmatic access to organized content. These instruments are proficient in retrieving tabular data, section titles, paragraphs, and metadata. PyMuPDF is recognized for its efficiency and ability to render images, whereas pdfplumber is superior in extracting information from organized PDF tables. The python-docx library offers detailed management of Word documents, allowing for the extraction and formatting of text, headers, and styles. These parsers are mainly designed for digital text inputs and cannot interpret handwritten or scanned examination papers.

*C. LLMs for Information Extraction*

Recent advancements in Large Language Models (LLMs), including GPT-4, Claude, and Mistral, have enabled more sophisticated information extraction and semantic understanding. These models can process OCR-extracted or parsed text using prompt-based parsing, allowing them to identify relevant sections, extract answers, match content to rubrics, and interpret context-sensitive information. Unlike rule-based systems, LLMs generalize across varying input structures and styles, making them suitable for evaluating descriptive answers. Integrating LLMs improves the system's ability to handle synonyms, paraphrased responses, and free-form writing, leading to more human-like, consistent evaluation.

Table 1: Comparison of OCR Tools, Document parsers, and LLMs

| Category | Tool/ Model | Strengths | Limitations |
| --- | --- | --- | --- |
| OCR Tools | Tesseract | Open Source, supports multiple language, widely used | Struggles with cursive handwriting, poor layout detection |
| | EasyOCR | Better support for handwritten and cursive text | Lower accuracy on complex layouts, less customizable |
| | Google Vision API | Better support for handwritten and cursive text | Paid API, internet required |
| Document Parsers | PyMuPdf | Fast rendering, text+image extraction | Limited section-wise structure parsing |
| | pdfplumber | Good table extraction and PDF layout parsing | Slow on large files, lacks formatting detection |
| | python-docx | Extracts paragraphs, metadata from .docx files | Only works with .docx, not scanned documents |
| LLMs | GPT-4 | Handles semantic similarity, paraphrasing | Needs careful prompting, computationally heavy |
| | Mistral | Lightweight, efficient open model for extraction | Less accurate on domain-specific tasks |
| | Claude | Strong reasoning and linguistic accuracy | Not open-source, limited availability |

IV. ARCHITECTURE

The proposed system structure for retrieving text from documents includes a multi-phase processing pipeline aimed at managing both scanned and digital document inputs, as shown in Fig. 1. This pipeline seeks to automate the transformation of diverse document formats into organized, machine-readable information, utilizing a blend of conventional preprocessing methods, OCR/digital parsing technologies, and sophisticated language models. The system accepts entries in PDF or image formats, including scanned handwritten or printed materials alongside digitally created files. This adaptability guarantees relevance across various document types.

1. *Preprocessing:* Input documents receive customized preprocessing according to their source. For scanned documents, techniques like binarization and skew correction are utilized to enhance text clarity and improve OCR accuracy. To prepare and clean the data before extraction, parsing libraries and noise removal algorithms are employed for digital documents.
2. *Text Extraction:* Two complementary approaches are utilized for retrieving unprocessed textual information. OCR engines, such as Tesseract, DocTR, and Google Vision, are employed to transform visual data from scanned images into text. Simultaneously, digital document parsers like pdfplumber and PyMuPDF obtain text and metadata directly from PDFs, allowing for effective handling of digitally created documents. The gathered data from both OCR and parsing components is combined into unstructured raw text, which acts as the input for the following semantic processing.
3. *LLM Processing:* Advanced Large Language Models (LLMs), including GPT-4 and Mistral, are subsequently utilized on the raw text via prompt-based methods to carry out focused information extraction. These models produce organized outputs, usually in JSON format, recognizing key-value pairs like names, dates, and addresses.
4. *Structured Output:* The concluding phase generates structured, machine-accessible data from the LLM output, allowing for efficient downstream applications like automated evaluation, indexing, or database retention.

This pipeline enables strong and automated comprehension of documents by combining classic image processing, cutting-edge OCR, and modern NLP techniques into a single framework.

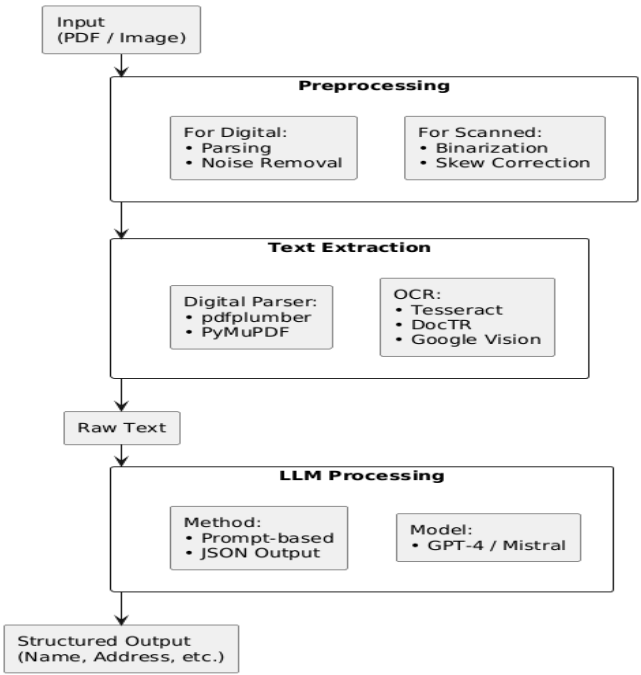

Fig. 1: Overview of the Document Text Extraction Pipeline

V. EXPERIMENTAL SETUP AND EVALUATION

A. DESCRIPTION OF THE DATASET

To thoroughly assess the efficacy of the suggested system, a mix of publicly accessible and proprietary datasets, generally classified into scanned and digital document categories.
The scanned documents are as follows:

1. *FUNSD (Form Understanding in Noisy Scanned Documents):* This dataset includes scanned forms marked with key-value pairs, facilitating the assessment of form comprehension in noisy environments.

2. *SROIE (Scanned Receipts OCR and Information Extraction):* Comprises scanned receipt images labeled with attributes like date, total amount, and vendor name, aiding in the evaluation of information extraction.

3. *Custom Invoice Dataset:* An exclusive compilation of scanned invoice images that have been meticulously labeled for optical character recognition (OCR) precision and key-value pair extraction efficacy.

The digital documents are as follows:

1. *Government Documents:* A variety of formatted PDF files such as tax forms, identification requests, and additional administrative templates.

2. *Research Papers:* Scientific articles created digitally in PDF format, tagged with structural metadata like title, author list, and abstract, valuable for assessing document layout comprehension.

The varied dataset structure allows for thorough evaluation of the system's effectiveness across various document types, including both degraded, scanned images and clear, native digital files.

B. OCR Evaluation

The Optical Character Recognition (OCR) component plays a critical role in processing scanned documents by converting image-based text into machine-readable formats. To assess OCR effectiveness, we conducted a comparative analysis of three OCR engines: Tesseract (v4), DocTR, and the Google Vision API. The evaluation was based on support for various formats, recognition accuracy, multilingual capabilities, and layout understanding. The results reveal notable differences in accuracy depending on document complexity, with the Google Vision API outperforming others in handling intricate layouts. These findings provide valuable insights into the strengths and limitations of each OCR engine for diverse document processing tasks.

Table 2: Comparison of OCR Engines

| OCR Tool | Format Support | Accuracy | Multi-language Support | Layout Awareness |
|---|---|---|---|---|
| Tesseract (v4) | Images only | ~85% | Yes | No |
| DocTR | PDF/Images | ~91% | Limited | Yes |
| Google Vision API | All formats | ~94% | Yes | Yes |

*C. Parsing Digital Documents*

In contrast to scanned handwritten documents that require OCR-based preprocessing, digitally created files—like government forms, research papers, and invoices in PDF and DOCX formats—have embedded text layers. This allows for accurate and straightforward extraction of textual and structural data through document parsers, eliminating the interference often linked to image-based processing.

*1) Instruments Employed:* To analyze and retrieve data from digitally generated documents, the subsequent libraries and tools are utilized:

- pdfplumber: Aids in retrieving text from PDFs, including table data and formatting specifics
- PyMuPDF (fitz): Provides detailed access to page content, such as metadata, embedded fonts, and images, with excellent fidelity.
- python-docx: Utilized for retrieving text and formatting details from DOCX documents.

*2) Fidelity of Extraction:* Digital documents facilitate accurate data extraction because of their structural features:

- Text Layer Accessibility: In contrast to raster images, digital PDFs retain text as character objects, which greatly improves extraction precision and removes errors caused by OCR.
- Maintained Structural Metadata: Internal components such as paragraphs, headings, and tables improve segmentation and contextual understanding.
- Encoding Consistency: In digital documents, font and character encoding are standardized, reducing the risk of interpretation mistakes.

Consequently, the confidence in extracting data from digital documents stays steadily near 100%.

*D. Key-Value Extraction Using LLMs*

Once the raw text is obtained, Large Language Models (LLMs) like GPT-4 and Mistral are utilized to convert unstructured text into organized key-value pairs by means of prompt engineering. This method is especially successful in retrieving details like Name, Date of Birth, Address, and Total Amount from different types of documents.Evaluates the model's capability to generate structured outputs that adhere to a specified JSON schema

VI. RESULT AND ANALYSIS

This part offers a thorough assessment of the suggested system focusing on various document processing activities, such as OCR accuracy, digital parsing accuracy, and structured information extraction utilizing LLMs.

*A. Comparison of OCR Engine Performance*

A comparative study was performed on three OCR engines: Tesseract (v4), DocTR, and Google Vision API. The assessment took into account recognition precision, multilingual support, and layout awareness.

- The Google Vision API exhibited the greatest accuracy, reaching an average of around 94%, and it also provided strong assistance for multilingual text and intricate layout designs.
- DocTR attained an impressive accuracy of around 91%, working well on both images and PDF types, with a fair level of layout recognition.
- Tesseract (v4) demonstrated a relatively lower accuracy of around 85% and did not offer native support for understanding intricate layouts.

*B. Accuracy of Digital Document Parsing*

For digitally created documents, tools such as pdfplumber, PyMuPDF, and python-docx were utilized, achieving nearly perfect extraction accuracy (~100%). This high precision is attributed to the inherent properties of digital files, including the preservation of text as discrete, selectable elements rather than images, which allows for exact retrieval. Additionally, these files retain important structural metadata—such as headings, sections, and tables—that aids in better segmentation and understanding of the document's content. Consistent character and font encoding throughout the files further reduces parsing errors, resulting in highly reliable and accurate text extraction.

*C. Extraction of Key-Value Pairs with LLMs*

After raw text extraction, Large Language Models (LLMs) like GPT-4 and Mistral were utilized to transform unstructured text into organized key-value pairs through prompt engineering. The models effectively extracted vital details like Name, Date of Birth, Address, and Total Amount from different types of documents.

The LLM-driven method achieved great accuracy with very few false positives, successfully producing results that met a specified JSON format. This validates the usefulness of LLMs for understanding documents at the semantic level.

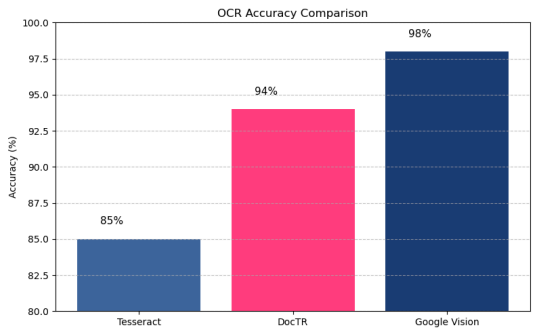

Fig 2. OCR Accuracy Comparison

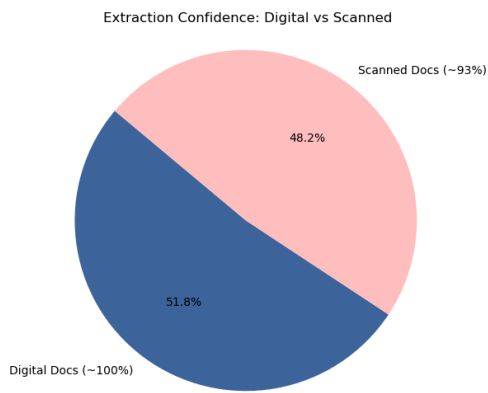

Fig 3. Extraction Confidence: Digital vs Scanned

## VII. CONCLUSION AND FUTURE SCOPE

This study introduces a hybrid system designed for the automated extraction and assessment of textual data from both scanned handwritten and digitally produced documents, merging Optical Character Recognition (OCR) approaches utilizing Large Language Models (LLMs) for understanding semantics and generating structured outputs.

Using preprocessing methods like binarization and skew correction on scanned inputs, along with OCR tools such as Tesseract, DocTR, and Google Vision, reliable text can be extracted from noisy handwritten documents. In digital formats, utilities like pdfplumber and PyMuPDF guarantee almost flawless text extraction because of the presence of embedded character information.

The concluding assessment module, powered by prompt-based LLMs like GPT-4 and Mistral, facilitates precise extraction of key-value pairs, scoring, and generation of structured responses. The system shows robust performance on datasets like FUNSD, SROIE, and personalized academic forms, assessed through metrics including word-level OCR accuracy, layout identification, and F1-score for extracting fields.

Although the suggested system yields encouraging outcomes, there are multiple opportunities for future improvements:

*A. Handwriting Style Modification:*

Utilizing deep learning-driven handwriting recognition models that are trained on specific domain scripts can enhance precision for cursive and artistic writing.

*B. Multilingual Assistance:*

Improving OCR and LLM elements to accommodate regional and multilingual documents can expand usability across various regions.

*C. Real-Time Feedback Mechanism:*

Incorporating the model into online or mobile platforms to enable immediate scanning, evaluation, and feedback for students and assessors.

ACKNOWLEDGMENT

The author gratefully acknowledges Mrs. Rekha B. S., mentor and guide at RV College of Engineering, for her essential support, motivation, and insights that influenced the course and caliber of this study. Thanks are also given to the faculty and staff of the Information Science and Engineering Department for their assistance and resources. The efforts of the researchers and developers involved in the cited papers and tools in OCR, NLP, and document understanding are sincerely valued, as their contributions greatly enhanced and informed this research